\documentclass[runningheads]{llncs}

\usepackage[T1]{fontenc}
\usepackage{amsmath,amsfonts}
\usepackage{graphicx}
\usepackage{booktabs}
\usepackage{multirow}
\usepackage{verbatim}

\begin{document}

\title{SurgSkill-Bench: A Benchmark for Multimodal Surgical Skill Assessment}

% Added for anonymized MICCAI 2025 submission
% \author{Anonymized Authors}
\author{
Chaohui Dang\inst{1} \and
Zheheng Jiang\inst{2} \and
James Glasbey\inst{3} \and
David Luke\inst{4} \and
Theodoros Arvanitis\inst{1} \and
Le Zhang\inst{1}\thanks{Corresponding author}
}
%index{Dang, Chaohui}
%index{Jiang, Zheheng}
%index{Glasbey, James}
%index{Luke, David}
%index{Arvanitis, Theodoros}
%index{Zhang, Le}

\authorrunning{Chaohui Dang et al.}

\institute{
Department of Electronic, Electrical and Systems Engineering,
School of Engineering,
University of Birmingham, Birmingham, UK
\and
School of Computing and Mathematical Science,
University of Leicester, Leicester, UK
\and
Department of Applied Health Sciences,
College of Medicine and Health,
University of Birmingham, Birmingham, UK
\and
Surgical Division, University Hospitals North Midlands, Stoke-on-Trent, UK
\\
\email{l.zhang.16@bham.ac.uk}
}
\maketitle

\begin{abstract}
Objective assessment of surgical technical skill is important for surgical training and structured feedback, but current workflows remain dependent on labor-intensive expert review. Existing automated approaches primarily focus on visual inputs and provide limited support for jointly studying operative performance, structured skill scores, and evaluator feedback. We introduce SurgSkill-Bench, an initial video-score-text benchmark-style dataset containing 214 surgical training simulation videos, six-dimensional OSATS scores, and expert free-text comments. We define two evaluation settings: video-only OSATS prediction for automated assessment and post hoc expert-comment-assisted prediction, where evaluator comments are available as auxiliary information. We provide controlled baseline experiments using representative frozen visual backbones, content-adaptive key-frame sampling, and a simple video-text co-attention fusion module. Under internal video-level validation, content-adaptive sampling improves video-only performance in this dataset, while evaluator comments provide additional score-related signal in the assisted setting. The best mean AUROC reaches 0.88 under dataset-specific median dichotomization. We further discuss evaluation constraints related to dataset scale, metadata completeness, and the interpretation of comment-assisted prediction. \textit{Code will be released publicly at a later date.}
\end{abstract}

\section{Introduction}
\label{sec:Introduction}

Objective assessment of surgical technical skill is central to surgical training, credentialing, and quality assurance. The Objective Structured Assessment of Technical Skill (OSATS) is a widely adopted standard in which expert raters score procedures across multiple dimensions using structured rubrics \cite{martin1997osats,niitsu2013osats,birkmeyer2013skill}. Despite its clinical relevance, OSATS requires manual expert review, making it labor-intensive, difficult to scale, and susceptible to inter-rater variability and subjective bias \cite{stulberg2020skillsoutcomes,gawad2019irr}. Although operative videos are increasingly used for education and assessment, recent surveys highlight the lack of standardized datasets and evaluation protocols for automated surgical skill assessment \cite{dick2024automatedvideo,lam2022mlskills}. Importantly, expert assessment in clinical practice often includes qualitative feedback explaining assigned scores, yet this semantic information is less commonly modeled together with operative video and structured skill scores \cite{peisl2024noise,delouche2024hrv}.

Advances in computer vision have enabled analysis of surgical video, including phase recognition, tool usage, and outcome prediction \cite{liu2021unifiedskills}. Beyond surgical video analysis, generative learning has been applied to standard-plane synthesis and missing-slice imputation in cardiac MRI, joint restoration of multiple image degradations, and diffusion-based brain-tumor inpainting \cite{zhang2019unsupervised,zhang2019missing,zhang2022learning,tao2025diffkan}. However, transferring general-purpose video representations to surgical skill assessment remains challenging. Surgical training videos contain long low-information intervals, repeated movements, and subtle instrument–tissue interactions, which can dilute skill-relevant cues under uniform sampling \cite{peisl2024noise,avellino2021summarization}. In addition, technical skill assessment is not based solely on visual appearance: expert raters often combine observed motion, tissue handling, instrument use, procedural flow, and qualitative reasoning when assigning OSATS scores. Existing automated studies have made progress in video-based assessment, but few benchmark settings explicitly connect operative video, structured multi-dimensional skill scores, and evaluator free-text comments within a unified evaluation protocol.

As an initial step toward studying these issues, we introduce SurgSkill-Bench, a pilot video-score-text benchmark for surgical skill assessment. The benchmark contains surgical training videos, six-dimensional OSATS ratings, and expert free-text comments. The primary goal of this work is to establish a reproducible benchmark setting and baseline protocol, rather than to claim a new state-of-the-art architecture. Our contributions are fourfold: (1) We introduce an initial video-score-text dataset of 214 surgical training simulation clips with six-dimensional OSATS scores and evaluator free-text comments. (2) We define two evaluation settings: video-only OSATS prediction and post hoc expert-comment-assisted OSATS prediction. (3) We provide a controlled baseline suite using representative frozen visual backbones, shared regression heads, and content-adaptive sampling. (4) We provide an explicit evaluation analysis covering thresholding, comment-assisted prediction, metadata constraints, and future participant-level validation.

\section{Dataset \& Benchmark}
\label{sec:Benchmark}

\subsection{Data Collection and Composition}

SurgSkill-Bench contains 214 curated surgical training simulation video clips provided by physicians. Each clip is approximately one minute long and records the operative field during task execution. The current release is designed for postoperative surgical skill assessment rather than intraoperative decision support. Each clip is linked to six-dimensional OSATS rating events and a set of
available evaluator free-text comments. Available scores are aggregated
by dimension to construct video-level consensus targets. Input frames are resized to $224 \times 224$ during preprocessing. Because expert annotation of surgical skill is costly, the current release remains modest in scale. We use a video-level split to prevent direct clip overlap between training, validation, and test partitions. Specifically, 10\% of videos are held out as an independent test set, and the remaining 90\% are used for five-fold cross-validation. In each fold, all frames, scores, and comments derived from the same video are assigned to the same partition. Model selection is performed using the validation fold only, and the held-out test set is not used for hyperparameter selection. The reported mean and standard deviation are computed across the five video-level folds unless otherwise stated. Because participant- and session-level metadata are incomplete, the current results should be interpreted as internal video-level benchmark performance rather than participant-level generalization.

\subsection{Annotations: OSATS and Expert Comments}

Each video was assessed by two or more expert expert raters using six OSATS dimensions: Respect for Tissue, Time \& Motion, Instrument Handling, Flow of Operation, Knowledge of Instruments, and Overall Score. Each dimension was scored on a 1--5 ordinal scale. The regression target for each dimension is the mean rater score, treated as a continuous consensus label. In addition to numerical OSATS scores, raters provided free-text comments describing technical performance, including tissue handling, instrument use, flow, motion efficiency, and overall technical quality. Since the same evaluators provided both comments and scores, comment-assisted experiments may contain score-related information and are interpreted as post hoc assisted prediction rather than autonomous video-only assessment.

\subsection{Benchmark Tasks}

Models are trained to regress continuous consensus scores on a 1--5 scale across the six OSATS dimensions. We report MAE and MSE as the primary regression metrics because the prediction target is the averaged rater score for each dimension. For supplementary discrimination analysis, we compute AUROC after median-based binarization for each dimension: scores greater than the median are treated as high-score samples, and scores at or below the median are treated as low-score samples. These thresholds are dataset-specific and should not be interpreted as clinically validated competency cutoffs. 

\section{Methodology}
\label{sec:Methodology}
\begin{figure}[!t]
  \centering
  \includegraphics[width=1\linewidth]{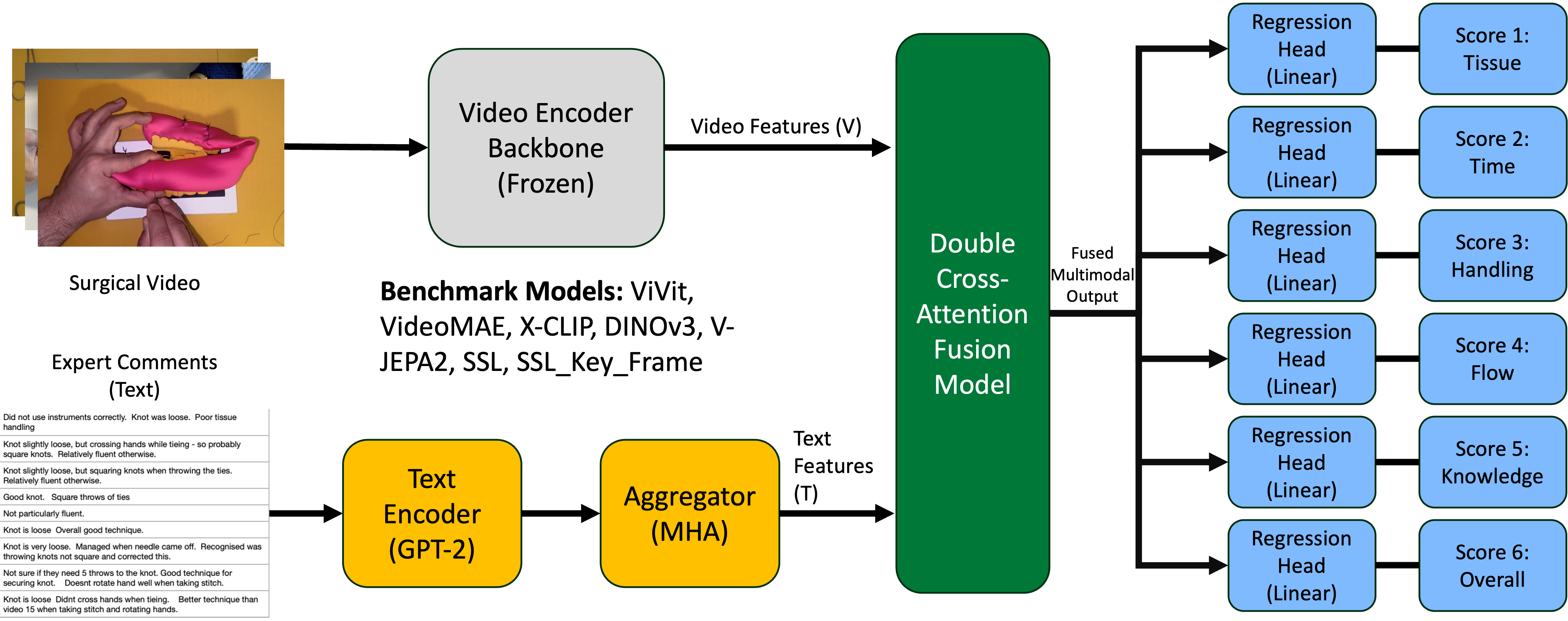}
  \caption{Overview of the expert-comment-assisted baseline framework. Frozen video encoder backbones and a frozen GPT-2 text encoder are used to extract visual and comment-derived features. A bidirectional co-attention module fuses the two representations, followed by regression heads to predict six OSATS scores. In the video-only setting, the text branch and fusion module are removed.}
  \label{fig:framework}
\end{figure}

To evaluate baseline algorithms on SurgSkill-Bench, we provide a modular framework for video-only and expert-comment-assisted OSATS prediction. As illustrated in Fig.~\ref{fig:framework}, the full assisted framework consists of three components: (1) a video representation module for extracting spatiotemporal features; (2) a text encoding module for representing evaluator comments; and (3) a fusion and regression head that combines visual and comment-derived features to predict six OSATS scores. For video-only experiments, the text branch and co-attention module are removed, and predictions are made from the video representation using the same type of regression heads.

\subsection{Content-Adaptive Key-Frame Extraction}

To reduce visual redundancy in surgical videos, we evaluate Content-Adaptive Key-Frame Extraction (CA-Frame) as a simple sampling baseline. A pre-trained InceptionV3 model extracts a 2048-dimensional feature vector for each frame \cite{szegedy2015inceptionv3}. Starting from the first frame as the reference, a new frame is selected as a key-frame when its cosine distance from the current reference exceeds a threshold $\tau=0.05$, after which the reference is updated. If too few key-frames are detected for the required input length, the data loader falls back to uniform sampling from the original clip. Otherwise, the selected key-frame pool is sampled in chronological order to form the fixed-length input sequence required by each backbone. Standard uniform sampling and CA-Frame use the same final input length, so CA-Frame changes only the temporal locations of selected frames. In the current dataset, CA-Frame selects an average of 80 frames from over 3000 original frames per clip. Since it uses ImageNet-pretrained InceptionV3 features, CA-Frame may miss subtle skill cues or fine-grained motion continuity.

\begin{figure}[!t]
  \centering
  \includegraphics[width=1\linewidth]{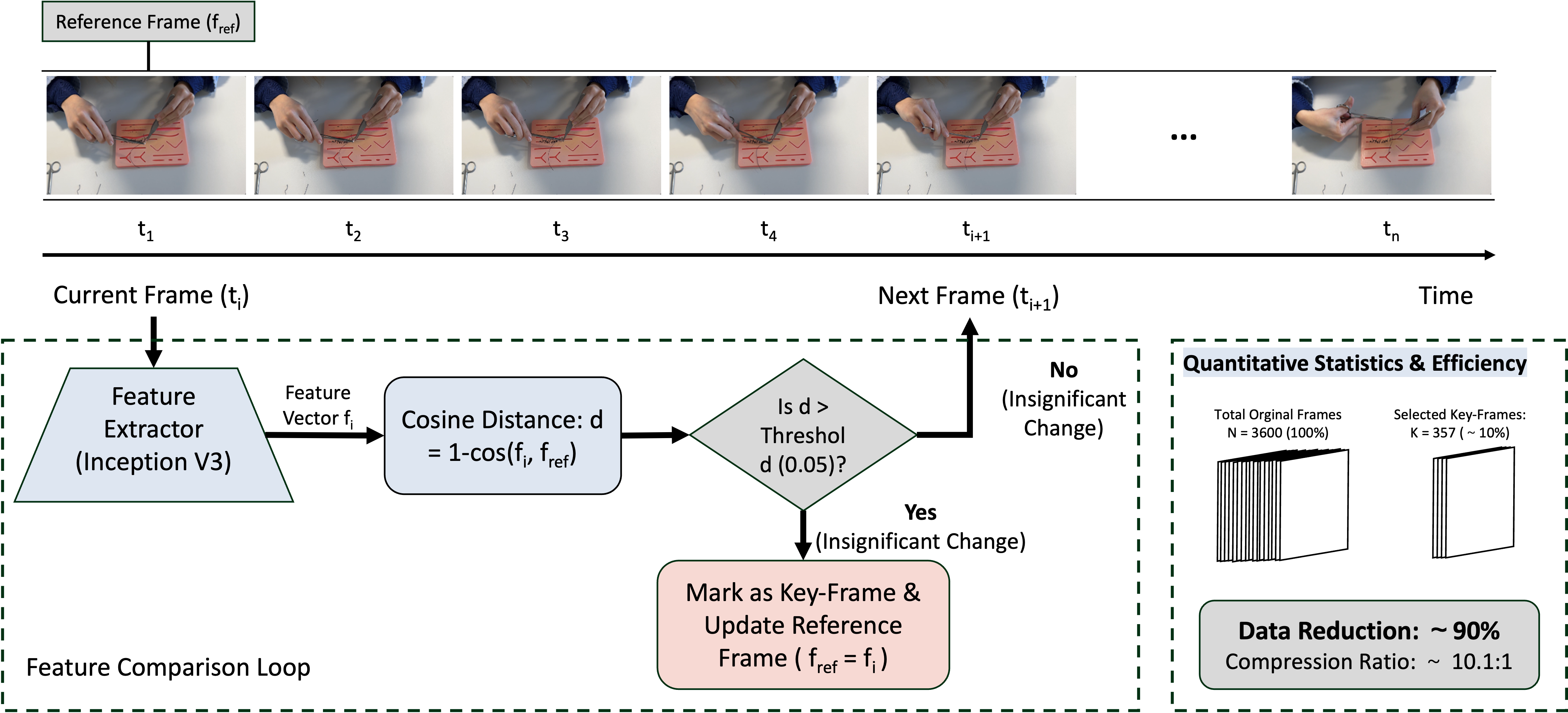}
  \caption{Content-Adaptive Key-Frame Extraction (CA-Frame) pipeline. InceptionV3 frame features and cosine distance to a reference frame are used to select key-frames, with uniform sampling applied when key-frames are sparse.}
  \label{fig:keyframe}
\end{figure}

\subsection{Expert-Comment-Assisted Fusion}

Expert comments are encoded using an off-the-shelf GPT-2 text encoder \cite{radford2019language}. Token-level features are pooled into comment embeddings, and rater comments are aggregated with multi-head attention to obtain a text representation $f_t$. Missing comments are represented by a neutral placeholder. The GPT-2 text encoder is kept frozen in all benchmark experiments. Only the text projection layer, comment aggregation module, fusion block, and regression heads are trained.

The co-attention module is used as a controlled fusion baseline rather than a novel architectural contribution. Let $F_v \in \mathbb{R}^{T \times d}$ denote the projected video feature sequence and let $F_t \in \mathbb{R}^{M \times d}$ denote the projected comment feature sequence, where $T$ is the number of visual tokens and $M$ is the number of available comment embeddings for a video. The bidirectional attention is computed as:

\begin{align}
    A_{v \rightarrow t} &= \mathrm{MHA}(F_v W_v, F_t W_t, F_t W_t), \\
    A_{t \rightarrow v} &= \mathrm{MHA}(F_t W_t, F_v W_v, F_v W_v).
\end{align}

The attended features are combined with residual connections, layer normalization, temporal pooling, and a feed-forward fusion block. In video-only experiments, the model is trained and evaluated without comments. Therefore, comment-assisted results are reported separately from video-only results.

\subsection{Multi-Task Regression and Optimization}

The video-only or fused representation is passed to six independent regression heads corresponding to the OSATS dimensions. The model is trained with the average MSE loss over the six consensus OSATS targets:

\begin{equation}
    \mathcal{L}_{total} = \frac{1}{6} \sum_{k=1}^{6} || \hat{y}_k - y_k ||^2_2 .
\end{equation}

The visual backbone is frozen in benchmark runs, while the trainable projection layers, fusion module, and regression heads are optimized for six-dimensional OSATS prediction. Unless otherwise stated, both visual and textual pretrained encoders are frozen. We use regression rather than ordinal classification because the training targets are averaged rater scores, which produce continuous consensus labels. Ordinal modeling is left for future benchmark extensions.

\section{Experiments}
\label{sec:experiments}

\textbf{Implementation Details.}
All experiments are implemented in PyTorch on an NVIDIA A100 GPU. Input frames are resized to $224 \times 224$. Standard sampling uniformly samples frames from each clip, whereas CA-Frame first selects candidate key-frames and then samples them in chronological order to match the same final input length. The final input length is 16 frames for all video backbones; for DINOv3, 16 frames are encoded independently and temporally pooled. All pretrained visual and textual encoders are frozen, and only projection layers, fusion modules, and six regression heads are trained.

Models are optimized with AdamW using a learning rate of $1 \times 10^{-4}$, weight decay $1 \times 10^{-4}$, batch size 8, and gradient clipping with maximum norm 1.0. Training is performed for 200 epochs with model selection based on validation MAE. Mixed precision is used consistently across experiments. The projection and fusion hidden dimension is 256, and the co-attention module uses 4 attention heads. Results are reported across five video-level folds with fixed split seeds. The held-out test set is not used for hyperparameter selection.

\textbf{Baselines.}
We evaluate representative pretrained visual backbones, including ViViT, VideoMAE, DINOv3, V-JEPA 2, X-CLIP, and a Surgical SSL baseline. To ensure a controlled comparison, all pretrained visual backbones are used as frozen feature extractors unless otherwise stated, and all models are evaluated with comparable projection layers and six OSATS regression heads. For image-level encoders such as DINOv3, frame-level features are temporally pooled to obtain video-level representations. For video backbones such as ViViT, VideoMAE, and V-JEPA~2, backbone outputs are projected into a shared hidden space before regression or multimodal fusion. Although X-CLIP is originally a video-language model, we evaluate it as a frozen video feature extractor under the same downstream OSATS regression protocol to ensure comparability with the other backbones. The Surgical SSL baseline uses a surgical-domain encoder pretrained on over 840 hours of publicly available surgical videos from SurgVU \cite{zia2025surgvu}.The SSL encoder was pretrained with 16-frame clips resized to $224 \times 224$ for 200 epochs using AdamW and a masked video modeling objective. No SurgSkill-Bench videos or labels were used during SSL pretraining. In downstream experiments, the SSL encoder is frozen and evaluated under the same projection-head and OSATS regression protocol as the other backbones.

\textbf{Evaluation Metrics.}
Because OSATS scores are ordinal scores on a 1--5 scale, MAE and MSE are used as the primary regression metrics. They are computed for each OSATS dimension and then macro-averaged across the six dimensions. We additionally report quadratic weighted Cohen's kappa between rounded model predictions and rounded consensus labels as a secondary ordinal agreement statistic. For kappa calculation, continuous predictions are rounded to the nearest integer and clipped to the valid OSATS range of 1--5 before comparison with expert-derived target scores. Kappa is computed per dimension and then macro-averaged. Because the reference labels are averaged rater scores, this statistic reflects agreement with rounded consensus labels rather than agreement with individual expert raters, and should not be interpreted as human inter-rater reliability. For supplementary discrimination analysis, AUROC is computed after median-based binarization as described in Section~\ref{sec:Benchmark}. Since the median thresholds are dataset-specific and not clinically validated competency cutoffs, AUROC is interpreted as a secondary benchmark statistic. We also include a mean-score predictor as a naive baseline to contextualize MAE and MSE. This baseline predicts the training-set mean score for each OSATS dimension.

\textbf{Quantitative Results.}
Table~\ref{tab:video_only} reports video-only performance under standard frame extraction and CA-Frame extraction, and Table~\ref{tab:multimodal} reports the corresponding post hoc expert-comment-assisted setting. CA-Frame is associated with improved video-only performance for most backbones under the current internal video-level protocol; for example, VideoMAE improves from 0.57 to 0.86 AUROC, and ViViT improves from 0.55 to 0.85 AUROC. Evaluator comments improve several models in the post hoc assisted setting, particularly under standard frame extraction, but the benefit is not uniform across all backbones and metrics. Under key-frame inputs, ViViT and V-JEPA~2 reach the best mean AUROC of 0.88 under dataset-specific median dichotomization. These results support the assisted-prediction setting but should not be conflated with autonomous video-only assessment.

\begin{table*}[t]
  \centering
  \caption{Performance of video-only models under Standard Frame Extraction and Key-Frame Extraction (Mean $\pm$ STD).}
  \label{tab:video_only}
  \resizebox{\textwidth}{!}{%
    \begin{tabular}{l|cccc|cccc}
      \toprule
      \multirow{2}{*}{\textbf{Model}} &
      \multicolumn{4}{c|}{\textbf{Standard Frame Extraction}} &
      \multicolumn{4}{c}{\textbf{Key-Frame Extraction}} \\
      \cmidrule(lr){2-5} \cmidrule(lr){6-9}
      & MAE $\downarrow$ & MSE $\downarrow$ & Kappa $\uparrow$ & AUROC $\uparrow$
      & MAE $\downarrow$ & MSE $\downarrow$ & Kappa $\uparrow$ & AUROC $\uparrow$ \\
      \midrule
      VideoMAE
      & $0.51 \pm 0.03$& $0.41 \pm 0.02$& $ 0.56 \pm 0.06 $& 
      $0.57 \pm 0.03 $& $0.16 \pm 0.01$& $0.04\pm 0.01$& $ 0.85 \pm 0.07 $&$0.86 \pm 0.06 $\\
      ViViT
      & $0.47 \pm 0.02$& $0.33\pm 0.02$& $ 0.61 \pm 0.08 $&
      $0.55 \pm 0.04 $& $0.20 \pm 0.02$& $0.06 \pm 0.01$& $ 0.81 \pm 0.06 $&$0.85 \pm 0.07 $\\
      X-CLIP
      & $0.67 \pm 0.11$& $0.64 \pm 0.19$& $ 0.45 \pm 0.04 $&
      $0.58 \pm 0.04 $& $0.36 \pm 0.07$& $0.26 \pm 0.09$& $ 0.83 \pm 0.05 $&$0.83 \pm 0.06$\\
      DINOv3
      & $0.78 \pm 0.04$& $0.83 \pm 0.06$& $ 0.33 \pm 0.06 $&
      $0.61 \pm 0.05$& $0.60 \pm 0.07$& $0.52 \pm 0.09$& $ 0.63 \pm 0.06 $&$0.79 \pm 0.03 $\\
      V-JEPA~2
      & $0.48 \pm 0.00$& $0.40 \pm 0.00$& $ 0.49 \pm 0.07 $&
      $0.56 \pm 0.06$& $0.15 \pm 0.01$& $0.05 \pm 0.01$& $ 0.87 \pm 0.05 $&$0.85 \pm 0.04 $\\
      SSL
      & $0.72 \pm 0.06$& $0.74 \pm 0.09$& $ 0.34 \pm 0.08 $&
      $0.66 \pm 0.05$& $0.34 \pm 0.00$& $0.30 \pm 0.00$& $ 0.84 \pm 0.07 $&$0.83 \pm 0.04 $\\
      \bottomrule
    \end{tabular}%
  }
\end{table*}

\begin{table*}[t]
  \centering
  \caption{Performance of post hoc expert-comment-assisted models under Standard Frame Extraction and Key-Frame Extraction (Mean $\pm$ STD). Expert comments are evaluator-provided feedback and may contain score-related information.}
  \label{tab:multimodal}
  \resizebox{\textwidth}{!}{%
    \begin{tabular}{l|cccc|cccc}
      \toprule
      \multirow{2}{*}{\textbf{Model}} &
      \multicolumn{4}{c|}{\textbf{Standard Frame Extraction}} &
      \multicolumn{4}{c}{\textbf{Key-Frame Extraction}} \\
      \cmidrule(lr){2-5} \cmidrule(lr){6-9}
      & MAE $\downarrow$ & MSE $\downarrow$ & Kappa $\uparrow$ & AUROC $\uparrow$
      & MAE $\downarrow$ & MSE $\downarrow$ & Kappa $\uparrow$ & AUROC $\uparrow$ \\
      \midrule
      VideoMAE
      & $0.76 \pm 0.01$& $0.80 \pm 0.01$& $ 0.36 \pm 0.05 $&
      $0.68 \pm 0.05 $& $0.14 \pm 0.02$& $0.03 \pm 0.01$& $ 0.84 \pm 0.07 $&$0.87 \pm 0.05$\\
      ViViT
      & $0.34 \pm 0.03$& $0.18 \pm 0.03$& $ 0.83 \pm 0.05 $&
      $0.81 \pm 0.04$& $0.13 \pm 0.02$& $0.03 \pm 0.01$& $ 0.85 \pm 0.08 $&$0.88 \pm 0.04$\\
      X-CLIP
      & $0.44 \pm 0.03$& $0.29 \pm 0.05$& $ 0.75 \pm 0.08$&
      $0.78 \pm 0.03$& $0.18 \pm 0.01$& $0.06 \pm 0.01$& $ 0.86\pm 0.07 $&$0.86 \pm 0.04$\\
      DINOv3
      & $0.59 \pm 0.02$& $0.49 \pm 0.02$& $ 0.34 \pm 0.06 $&
      $0.75 \pm 0.02$& $0.27 \pm 0.02$& $0.12 \pm 0.02$& $ 0.79 \pm 0.09 $&$0.84 \pm 0.02$\\
      V-JEPA~2& $0.56 \pm 0.06$& $0.50 \pm 0.07$& $ 0.60 \pm 0.08 $&
      $0.75 \pm 0.03$& $0.13 \pm 0.02$& $0.03 \pm 0.01$& $ 0.85 \pm 0.08 $&$0.88 \pm 0.04$\\
      SSL 
      & $0.73 \pm 0.03$& $0.74 \pm 0.04$& $ 0.35 \pm 0.07 $&
      $0.70 \pm 0.02$& $0.16 \pm 0.02$& $0.04 \pm 0.01$& $ 0.85 \pm 0.07 $&$0.87 \pm 0.03$\\
      \bottomrule
    \end{tabular}%
  }
\end{table*}

\begin{figure}[!t]
  \centering
  \includegraphics[width=1\linewidth]{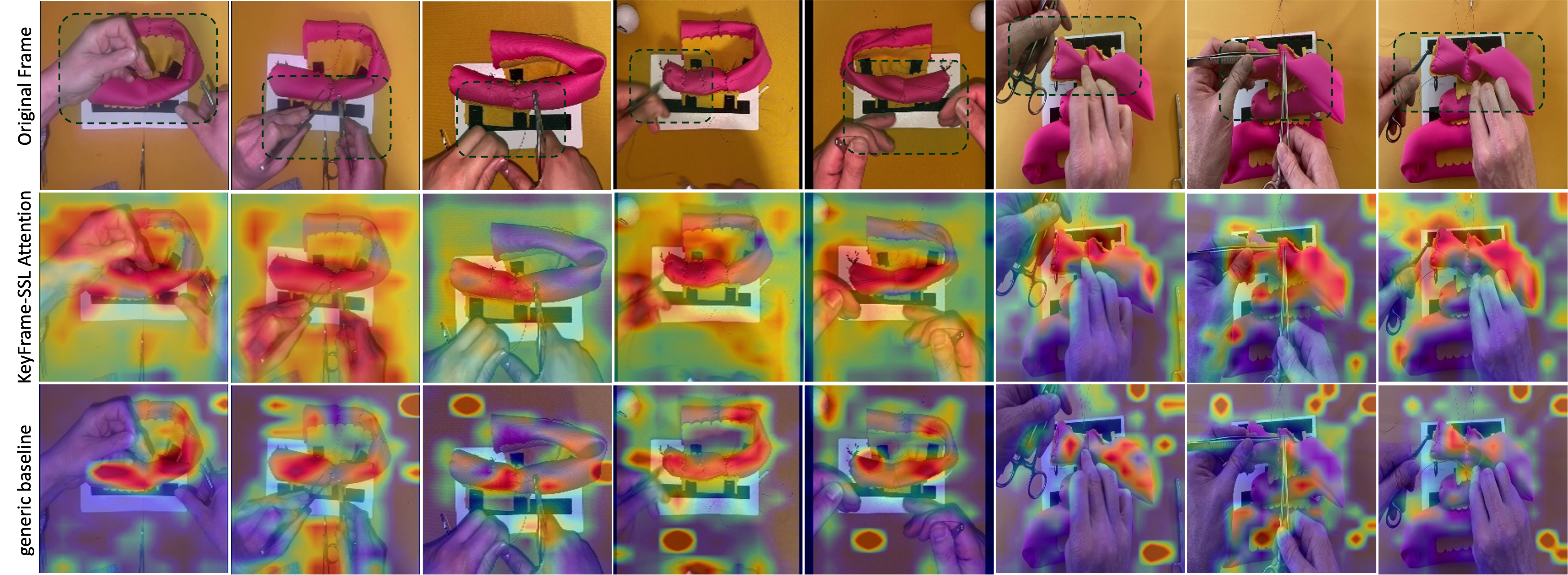}
  \caption{Qualitative attention visualization. The first row shows original surgical frames with manual annotations of instrument, hand, and tissue interaction regions. The second row shows attention maps from the CA-Frame-guided surgical SSL encoder, and the third row shows attention maps from a generic visual baseline.}
  \label{fig:heatmap}
\end{figure}

\textbf{Qualitative Visualization.}
Fig.~\ref{fig:heatmap} shows self-attention heatmaps from validation samples. In these selected examples, the CA-Frame-guided SSL encoder visually appears to produce more concentrated responses around regions containing instruments, hands, and tissue interactions, whereas the generic visual baseline appears more diffuse. These visualizations are qualitative examples only and should not be interpreted as quantitative evidence of interpretability or causal proof of clinically meaningful reasoning.

\section{Conclusion}
\label{sec:Conclusion}

We introduced SurgSkill-Bench, an initial video-score-text benchmark for surgical skill assessment that combines surgical training videos, six-dimensional OSATS scores, and evaluator free-text comments. We defined separate video-only and post hoc expert-comment-assisted settings and provided controlled baselines using frozen visual backbones, content-adaptive sampling, and video-text fusion. Under internal video-level validation, CA-Frame was associated with improved video-only performance, while evaluator comments provided additional score-related signal in the assisted setting. These findings establish a benchmark baseline, while remaining constrained by dataset scale, incomplete participant metadata, dataset-specific AUROC thresholds, and the assisted nature of comment-based prediction. Future work will expand the dataset, add participant-level validation, include text-leakage controls, and analyze human agreement.

\section*{Disclosure of Interests}
The authors have no competing interests to declare.

\bibliographystyle{splncs04}
\bibliography{reference}

@article{niitsu2013osats,
  title   = {Using the Objective Structured Assessment of Technical Skills (OSATS) Global Rating Scale to Evaluate the Skills of Surgical Trainees in the Operating Room},
  author  = {Niitsu, Hiroaki and Hirabayashi, Naoki and Yoshimitsu, Masanori and Mimura, Takeshi and Taomoto, Junya and Sugiyama, Yoichi and Murakami, Shigeru and Saeki, Shuji and Mukaida, Hidenori and Takiyama, Wataru},
  journal = {Surgery Today},
  year    = {2013},
  volume  = {43},
  number  = {3},
  pages   = {271--275},
  doi     = {10.1007/s00595-012-0313-7},
  url     = {https://link.springer.com/article/10.1007/s00595-012-0313-7}
}

@misc{zia2025surgvu,
  title         = {Surgical Visual Understanding (SurgVU) Dataset},
  author        = {Zia, Aneeq and Berniker, Max and Nespolo, Rogerio and Perreault, Conor and Wang, Ziheng and Mueller, Benjamin and Schmidt, Ryan and Bhattacharyya, Kiran and Liu, Xi and Jarc, Anthony},
  year          = {2025},
  eprint        = {2501.09209},
  archivePrefix = {arXiv},
  primaryClass  = {cs.CV},
  doi           = {10.48550/arXiv.2501.09209},
  url           = {https://arxiv.org/abs/2501.09209}
}

@misc{szegedy2015inceptionv3,
  title         = {Rethinking the Inception Architecture for Computer Vision},
  author        = {Szegedy, Christian and Vanhoucke, Vincent and Ioffe, Sergey and Shlens, Jonathon and Wojna, Zbigniew},
  year          = {2015},
  eprint        = {1512.00567},
  archivePrefix = {arXiv},
  primaryClass  = {cs.CV},
  doi           = {10.48550/arXiv.1512.00567},
  url           = {https://arxiv.org/abs/1512.00567}
}

@article{martin1997osats,
  title   = {Objective Structured Assessment of Technical Skill (OSATS) for Surgical Residents},
  author  = {Martin, J. A. and Regehr, G. and Reznick, R. and Macrae, H. and Murnaghan, J. and Hutchison, C. and Brown, M.},
  journal = {British Journal of Surgery},
  year    = {1997},
  volume  = {84},
  number  = {2},
  pages   = {273--278},
  doi     = {10.1046/j.1365-2168.1997.02502.x},
  url     = {https://doi.org/10.1046/j.1365-2168.1997.02502.x}
}

@article{delouche2024hrv,
  title   = {Heart Rate Variability as a Dynamic Marker of Surgeons’ Stress During Vascular Surgery},
  author  = {De Louche, Calvin D. and Mandal, Manish and Fernandes, Lee and Lawson, Jason and Bicknell, Colin D. and Pouncey, Anna L.},
  journal = {BJS Open},
  year    = {2024},
  volume  = {8},
  number  = {5},
  pages   = {zrae097},
  doi     = {10.1093/bjsopen/zrae097},
  url     = {https://doi.org/10.1093/bjsopen/zrae097}
}

@article{peisl2024noise,
  title   = {Noise in the Operating Room Coincides with Surgical Difficulty},
  author  = {Peisl, Sarah and S{\'a}nchez-Taltavull, Daniel and Guillen-Ramirez, Hugo and Tschan, Franziska and Semmer, Norbert K. and H{\"u}bner, Martin and Demartines, Nicolas and Wrann, Simon G. and Gutknecht, Stefan and Weber, Markus and Candinas, Daniel and Beldi, Guido and Keller, Sandra},
  journal = {BJS Open},
  year    = {2024},
  volume  = {8},
  number  = {5},
  pages   = {zrae098},
  doi     = {10.1093/bjsopen/zrae098},
  url     = {https://doi.org/10.1093/bjsopen/zrae098}
}

@article{dick2024automatedvideo,
  title   = {Automated Analysis of Operative Video in Surgical Training: Scoping Review},
  author  = {Dick, Lachlan and Boyle, Connor P. and Skipworth, Richard J. E. and Smink, Douglas S. and Tallentire, Victoria Ruth and Yule, Steven},
  journal = {BJS Open},
  year    = {2024},
  volume  = {8},
  number  = {5},
  pages   = {zrae124},
  doi     = {10.1093/bjsopen/zrae124},
  url     = {https://doi.org/10.1093/bjsopen/zrae124}
}

@article{birkmeyer2013skill,
  title   = {Surgical Skill and Complication Rates After Bariatric Surgery},
  author  = {Birkmeyer, John D. and Finks, Jonathan F. and O'Reilly, Amanda and Oerline, Mary and Carlin, Arthur M. and Nunn, Andre R. and Dimick, Justin and Banerjee, Mousumi and Birkmeyer, Nancy J. O.},
  journal = {New England Journal of Medicine},
  year    = {2013},
  volume  = {369},
  number  = {15},
  pages   = {1434--1442},
  doi     = {10.1056/NEJMsa1300625},
  url     = {https://doi.org/10.1056/NEJMsa1300625}
}

@article{stulberg2020skillsoutcomes,
  title   = {Association Between Surgeon Technical Skills and Patient Outcomes},
  author  = {Stulberg, Jonah J. and Huang, Reiping and Kreutzer, Lindsey and Ban, Kristen and Champagne, Bradley J. and Steele, Scott R. and Johnson, Julie K. and Holl, Jane L. and Greenberg, Caprice C. and Bilimoria, Karl Y.},
  journal = {JAMA Surgery},
  year    = {2020},
  volume  = {155},
  number  = {10},
  pages   = {960--968},
  doi     = {10.1001/jamasurg.2020.3007},
  url     = {https://doi.org/10.1001/jamasurg.2020.3007}
}

@article{gawad2019irr,
  title   = {The Inter-Rater Reliability of Technical Skills Assessment and Retention of Rater Training},
  author  = {Gawad, Nada and Fowler, Amanda and Mimeault, Richard and Raiche, Isabelle},
  journal = {Journal of Surgical Education},
  year    = {2019},
  volume  = {76},
  number  = {4},
  pages   = {1088--1093},
  doi     = {10.1016/j.jsurg.2019.01.001},
  url     = {https://www.sciencedirect.com/science/article/pii/S193172041830641X}
}

@article{lam2022mlskills,
  title   = {Machine Learning for Technical Skill Assessment in Surgery: A Systematic Review},
  author  = {Lam, K. and Chen, J. and Wang, Z. and others},
  journal = {npj Digital Medicine},
  year    = {2022},
  volume  = {5},
  pages   = {24},
  doi     = {10.1038/s41746-022-00566-0},
  url     = {https://doi.org/10.1038/s41746-022-00566-0}
}

@inproceedings{liu2021unifiedskills,
  title     = {Towards Unified Surgical Skill Assessment},
  author    = {Liu, Daochang and Li, Qiyue and Jiang, Tingting and Wang, Yizhou and Miao, Rulin and Shan, Fei and Li, Ziyu},
  booktitle = {Proceedings of the IEEE/CVF Conference on Computer Vision and Pattern Recognition (CVPR)},
  year      = {2021},
  pages     = {9522--9531},
  doi       = {10.48550/arXiv.2106.01035},
  url       = {https://arxiv.org/abs/2106.01035}
}

@article{avellino2021summarization,
  title   = {Surgical Video Summarization: Multifarious Uses, Summarization Process and Ad-Hoc Coordination},
  author  = {Avellino, Ignacio and Nozari, Sheida and Canlorbe, Geoffroy and Jansen, Yvonne},
  journal = {Proc. ACM Hum.-Comput. Interact.},
  year    = {2021},
  volume  = {5},
  number  = {CSCW1},
  articleno= {140},
  numpages = {23},
  doi     = {10.1145/3449214},
  url     = {https://doi.org/10.1145/3449214}
}

@misc{radford2019language,
  title={Language Models are Unsupervised Multitask Learners},
  author={Radford, Alec and Wu, Jeff and Child, Rewon and Luan, David and Amodei, Dario and Sutskever, Ilya},
  year={2019}
}

@inproceedings{zhang2019unsupervised,
  title={Unsupervised Standard Plane Synthesis in Population Cine MRI via Cycle-Consistent Adversarial Networks},
  author={Zhang, Le and Perea{\~n}ez, Marco and Bowles, Christopher and Piechnik, Stefan K and Neubauer, Stefan and Petersen, Steffen E and Frangi, Alejandro F},
  booktitle={(MICCAI 2019) International Conference on Medical Image Computing and Computer-Assisted Intervention},
  pages={660--668},
  year={2019},
  organization={Springer, Cham}
}

@inproceedings{zhang2019missing,
  title={Missing Slice Imputation in Population CMR Imaging via Conditional Generative Adversarial Nets},
  author={Zhang, Le and Perea{\~n}ez, Marco and Bowles, Christopher and Piechnik, Stefan and Neubauer, Stefan and Petersen, Steffen and Frangi, Alejandro},
  booktitle={(MICCAI 2019 Best Paper Finalist) International Conference on Medical Image Computing and Computer-Assisted Intervention},
  pages={651--659},
  year={2019},
  organization={Springer, Cham}
}

@article{zhang2022learning,
  title={Learning to restore multiple image degradations simultaneously},
  author={Zhang, Le and Bronik, Kevin and Papie{\.z}, Bart{\l}omiej W},
  journal={Pattern Recognition},
  volume={136},
  pages={109250},
  year={2022},
  publisher={Pergamon}
}

@inproceedings{tao2025diffkan,
  title={DiffKAN-Inpainting: KAN-based Diffusion model for brain tumor inpainting},
  author={Tao, Tianli and Wang, Ziyang and Zhang, Han and Arvanitis, Theodoros N and Zhang, Le},
  booktitle={Proceedings of IEEE International Symposium on Biomedical Imaging (ISBI)},
  year={2025}
}

\end{document}